\documentclass[conference]{IEEEtran}

\usepackage{amsmath,amssymb,amsfonts}
\usepackage{booktabs}
\usepackage{graphicx}
\usepackage{array}
\usepackage{float}
\usepackage{hyperref}
\usepackage{siunitx}

\usepackage{cite}

\title{BaziQA-Benchmark: Evaluating Symbolic and Temporally Compositional Reasoning in Large Language Models}

\author{
\IEEEauthorblockN{Jiangxi Chen}
\IEEEauthorblockA{
Shanghai Jiao Tong University\\
Shanghai, China\\
Email: jiangxi\_chen@163.com
}
\and
\IEEEauthorblockN{Qian Liu}
\IEEEauthorblockA{
Shanghai Zhinan Information Technology\\
Shanghai, China\\
Email: swaylq0913@gmail.com
}
}

\begin{document}
\maketitle

\maketitle

\begin{abstract}
We present \textsc{BaziQA-Benchmark}, a standardized benchmark for evaluating symbolic and temporally compositional reasoning in large language models.
The benchmark is derived from 200 professionally curated, multiple-choice problems from the Global Fortune-teller Competition (2021--2025), where each instance requires structured inference over a fixed symbolic chart and interacting temporal conditions.
Unlike anecdotal or prompt-driven evaluations, \textsc{BaziQA-Benchmark} enables objective scoring and controlled comparison across years, domains, and model families.
We evaluate contemporary language models under a multi-turn setting and analyze performance variation across temporal difficulty, reasoning domains, and inference protocols.
To further probe reasoning behavior, we introduce a lightweight Structured Reasoning Protocol that constrains inference order without adding domain knowledge.
Results show that models consistently outperform chance but remain far from saturation, exhibiting pronounced sensitivity to temporal composition and reasoning order, as well as systematic failures on precise temporal localization and multi-condition symbolic judgments.
\end{abstract}

% \begin{IEEEkeywords}
% symbolic reasoning, temporal reasoning, benchmark evaluation, large language models
% \end{IEEEkeywords}

\section{Introduction}
\label{sec:intro}

Large language models (LLMs) have achieved substantial progress on a variety of reasoning-oriented benchmarks, including mathematical problem solving, program synthesis, symbolic manipulation, and multi-hop question answering~\cite{shi2026reasoningtrees}.
Large-scale evaluations such as BIG-Bench and its challenging subsets demonstrate that modern models can internalize non-trivial inference patterns beyond surface-level text generation~\cite{srivastava2022bigbench,suzgun2022bbh}.
Complementary benchmarks based on academic examinations and knowledge-intensive reasoning further suggest that LLMs can generalize across heterogeneous reasoning formats~\cite{hendrycks2021mmlu,clark2018arc}.
However, most existing evaluations focus on domains with standardized semantics, formal operators, or well-defined symbolic languages.
Systematic assessment of LLM reasoning in \emph{structured but non-standard symbolic systems}, where inference depends on domain-specific interaction rules and heterogeneous symbolic effects, remains limited.

Bazi (Eight Characters) astrology provides a distinctive testbed for such reasoning.
It is a traditional East Asian symbolic system in which conclusions about life events and personal attributes are derived from a natal chart composed of Heavenly Stems and Earthly Branches, together with interacting temporal layers such as decadal and annual cycles.
Competition-grade Bazi problems require a constrained inference process: global chart assessment, prioritization of symbolic interactions, and mapping of dominant signals to concrete outcomes via predefined relational structures (e.g., Ten Gods and palaces).
As a result, Bazi reasoning more closely resembles multi-step symbolic inference under explicit constraints than free-form narrative generation.

For benchmarking purposes, a key challenge is that Bazi reasoning operates over a derived natal chart rather than raw birth information.
To isolate symbolic inference from calendrical conversion and chart derivation, \textsc{BaziQA-Benchmark} provides models with a pre-computed and consistently formatted chart context for each subject, and evaluates question answering conditioned on that fixed representation. This design enables controlled comparison across models and prompting strategies under identical symbolic inputs. Related work on long-context compression in other structured domains shows that preserving task-critical structure under constrained budgets is non-trivial, motivating standardized and consistent context representations in evaluation~\cite{shi2025longcodezip}.

Existing evaluations of LLMs on culturally grounded or metaphysical reasoning tasks remain fragmented.
Many studies rely on anecdotal examples, qualitative case analyses, or prompt-driven demonstrations, limiting reproducibility and comparability.
The most closely related BaZi-focused benchmark emphasizes persona consistency and character simulation under temporal settings~\cite{zheng2025bazi_character_sim}.
While valuable for studying generative behavior, such simulation-oriented tasks do not isolate symbolic inference accuracy under objective scoring.
To date, there is no standardized benchmark that evaluates professional-grade Bazi reasoning across years, domains, and model families using competition-grade questions and reproducible protocols.

In this work, we introduce \textbf{BaziQA-Benchmark}, a standardized evaluation suite constructed from 200 official problems of a global professional Bazi competition held between 2021 and 2025.
All questions are multiple-choice and professionally curated, enabling objective scoring.
The benchmark spans multiple application domains encountered in real-world Bazi practice, including marriage, career, health, wealth, personality, family relations, and annual fortune, and incorporates both static symbolic relations and temporally compositional reasoning.
Its design supports fine-grained analysis across competition years, reasoning domains, and interaction protocols under a unified evaluation setting.

Beyond benchmark construction, we examine the role of reasoning structure at inference time.
Prior work has shown that explicitly eliciting intermediate reasoning steps can improve performance on complex tasks~\cite{wei2022cot}, and that aggregation strategies such as self-consistency can reduce variability across reasoning trajectories~\cite{wang2023selfconsistency}.
Motivated by these findings, we introduce a three-stage \emph{Structured Reasoning Protocol} as an evaluation scaffold for Bazi inference.
The protocol decomposes reasoning into chart-level quantitative analysis, interaction severity grading, and domain-specific symbol-to-event mapping, without introducing additional domain knowledge or supervision.
This enables controlled analysis of how reasoning structure affects accuracy, stability, and failure patterns.

Through experiments across several contemporary model families and five competition years, we show that modern LLMs perform significantly above chance on professional Bazi problems, but remain far from saturation.
Performance varies substantially across domains, years, and inference protocols, and structured reasoning yields heterogeneous effects rather than uniform gains.
Overall, \textsc{BaziQA-Benchmark} provides a controlled and reproducible testbed for studying symbolic and temporally compositional reasoning, clarifying both the capabilities and current limitations of large language models in this class of domains. The benchmark data, evaluation scripts, and accompanying technical report are publicly released at \url{https://github.com/ChenJiangxi/BaziQA}.

\section{Benchmark Definition and Evaluation Protocol}
\label{sec:benchmark}

This section specifies the construction of \textsc{BaziQA-Benchmark} and the evaluation protocol adopted in this work.
The benchmark is designed to formalize professional Bazi astrology reasoning as a reproducible, objectively scorable task,
while preserving the symbolic structure and temporal dependencies encountered in real practice.

\subsection{Problem Source and Benchmark Scope}
\label{subsec:source}

\textsc{BaziQA-Benchmark} is constructed from official problems of a global professional Bazi competition held annually from 2021 to 2025.
All problems are authored and curated by experienced practitioners and used in competitive evaluation settings.

Each competition year consists of eight subjects (\emph{mingzhu}), where each subject corresponds to a fixed natal chart and is associated with five multiple-choice questions.
This results in 40 questions per year and 200 questions in total.
All questions are four-option multiple-choice (A/B/C/D), yielding a random-guess baseline accuracy of 25\%.

The benchmark is explicitly designed for objective scoring.
Official answer keys are provided for all questions, eliminating the need for post-hoc interpretation or human adjudication.
This property distinguishes \textsc{BaziQA-Benchmark} from open-ended narrative or simulation-based evaluations.

\subsection{Task Structure and Domain Coverage}
\label{subsec:task}

Each subject’s questions span multiple application domains commonly encountered in applied Bazi practice, including marriage, career, family relations, personality, health, wealth, education, and annual fortune.
Questions within a subject are ordered to approximately reflect increasing inferential complexity, ranging from static chart interpretation to multi-factor and temporally compositional reasoning.

The benchmark does not evaluate free-form explanation or narrative generation.
Instead, each question requires a discrete symbolic judgment grounded in domain-specific rules and precedence relations.
This formulation isolates reasoning correctness from stylistic fluency and enables controlled comparison across models and prompting protocols.

\subsection{Input Construction and Chart Representation}
\label{subsec:input}

In professional practice, Bazi reasoning is performed over a derived natal chart rather than raw birth information.
To isolate symbolic inference from calendrical conversion and chart derivation, \textsc{BaziQA-Benchmark} adopts a two-stage input construction process.

For each subject, the dataset provides a birth profile (e.g., date and time of birth and related attributes).
An external Bazi charting procedure is applied to compute the complete natal chart and its temporal expansions, including Heavenly Stems and Earthly Branches, Ten Gods relations, decadal luck cycles, and annual influences.
The resulting chart is stored in structured form and rendered into a fixed, human-readable textual template.

This formatted chart context is presented once at the beginning of each evaluation session and remains fixed while the model answers the five associated questions.
All models therefore receive identical symbolic information.
Personal identifiers are removed and replaced with anonymous subject IDs to avoid leakage of extraneous information.

\subsection{Multi-turn Evaluation Protocol}
\label{subsec:interaction}

For each subject, the model is first provided with the formatted natal chart context.
The five associated questions are then answered sequentially within a single conversational session.

This multi-turn design reflects realistic evaluation settings, where multiple judgments are made with respect to the same chart.
It also enables analysis of contextual accumulation effects, such as whether earlier answers implicitly condition later reasoning.
Models do not receive feedback on answer correctness during the session.

\subsection{Structured Reasoning Protocol as an Evaluation Scaffold}
\label{subsec:srp}

Bazi astrology reasoning involves multiple symbolic effects whose influences are heterogeneous and non-additive.
Correct inference therefore depends on identifying dominant interactions among competing factors rather than exhaustively enumerating all applicable rules.

To make this prioritization explicit, we introduce a \emph{Structured Reasoning Protocol} (SRP) as an evaluation scaffold.
The protocol constrains the reasoning process into a small number of ordered steps that reflect expert practice while remaining model-agnostic.

Given a fixed natal chart and a target question, inference under SRP proceeds as follows:
\begin{enumerate}[leftmargin=1.2cm]
\item \textbf{Quantitative Scan}: establish chart-level priors by assessing element balance, Day Master strength, and global structural patterns.
\item \textbf{Severity Grading}: identify and rank symbolic interactions according to their relative dominance under the current temporal context.
\item \textbf{Event Mapping}: map dominant symbolic signals to concrete outcomes using domain-specific rules, giving precedence to higher-severity interactions.
\end{enumerate}

The protocol does not introduce additional domain knowledge or supervision.
Its role is to constrain the order and completeness of inference steps, enabling controlled analysis of how reasoning structure affects accuracy, stability, and failure modes.
In this work, SRP is used exclusively at inference time and does not modify model parameters.

\subsection{Evaluation Settings and Aggregation}
\label{subsec:evaluation}

All models are evaluated without task-specific fine-tuning.
Unless otherwise stated, each model--year configuration is evaluated using multiple independent runs to account for stochasticity in decoding.

Accuracy is computed as the proportion of correctly selected answers.
For aggregate reporting, we compute macro-average accuracy across subjects, years, and domains.
To characterize robustness, we additionally report dispersion measures capturing run-to-run and year-to-year variability.

Statistical comparisons against the random-guess baseline are conducted using standard proportion tests.
Comparisons across prompting protocols and model families are reported using paired analyses where applicable.

\section{Benchmark Evaluation and Analysis}
\label{sec:evaluation}

This section presents the empirical evaluation of \textsc{BaziQA-Benchmark} under the benchmark protocol defined in Section~\ref{sec:benchmark}.
Rather than pursuing a new state of the art, the objective is to characterize how contemporary large language models behave under a structured symbolic reasoning benchmark with temporal composition and domain heterogeneity.
Accordingly, we analyze overall benchmark difficulty, year-wise variation, domain-wise performance profiles, sensitivity to inference-time prompting strategies, and cross-model agreement patterns.

Unless otherwise specified, all results are obtained under the primary multi-turn evaluation setting.
Macro-average accuracy is computed by first averaging within each competition year and then averaging across years, treating each year as an equal evaluation unit.
This aggregation strategy avoids overweighting any single competition set and reflects the intended use of the benchmark as a longitudinal evaluation suite rather than a single static test.

\subsection{Overall Benchmark Performance}
\label{subsec:overall_performance}

Table~\ref{tab:overall_main} reports macro-average accuracy over the five competition years (2021--2025).
Year-level 95\% confidence intervals are computed by treating the five yearly accuracies as independent observations and applying a $t$-interval ($n=5$).
The substantial overlap among confidence intervals indicates that differences in macro-average accuracy should be interpreted as descriptive ordering rather than statistically decisive separation at the benchmark scale.

\begin{table}[t]
\centering
\small
\caption{Macro-average accuracy over five competition years (Multi-turn).}
\label{tab:overall_main}
\setlength{\tabcolsep}{5pt}
\begin{tabular}{lcc}
\toprule
Model & Macro-Avg. (\%) & 95\% CI \\
\midrule
DeepSeek-Chat-V3 & 36.7 & [33.2, 40.2] \\
DeepSeek-R1      & 34.1 & [29.6, 38.6] \\
GPT-5.1-Chat     & 32.5 & [29.0, 36.0] \\
Gemini-2.5-Flash & 32.4 & [29.7, 35.1] \\
Gemini-3-Pro     & 32.1 & [26.6, 37.6] \\
\bottomrule
\end{tabular}
\end{table}

All evaluated models perform significantly above the 25\% random baseline (binomial test, $p < 10^{-7}$ for all models), confirming that \textsc{BaziQA-Benchmark} captures non-trivial symbolic structure rather than superficial pattern matching.
At the same time, absolute accuracy remains far from saturation, with even the strongest models answering fewer than half of the questions correctly.
This gap highlights the intrinsic difficulty of professional-grade Bazi reasoning, where multiple symbolic effects must be prioritized and composed under temporal constraints.

Notably, the relatively narrow spread between models suggests that no single architecture or training paradigm yields a decisive advantage at the benchmark level.
Instead, differences appear to arise from domain-specific inductive biases and inference strategies, which are further examined in subsequent analyses.

\subsection{Year-wise Performance and Temporal Difficulty}
\label{subsec:yearwise_performance}

\begin{figure*}[t]
\centering
\includegraphics[width=0.8\textwidth]{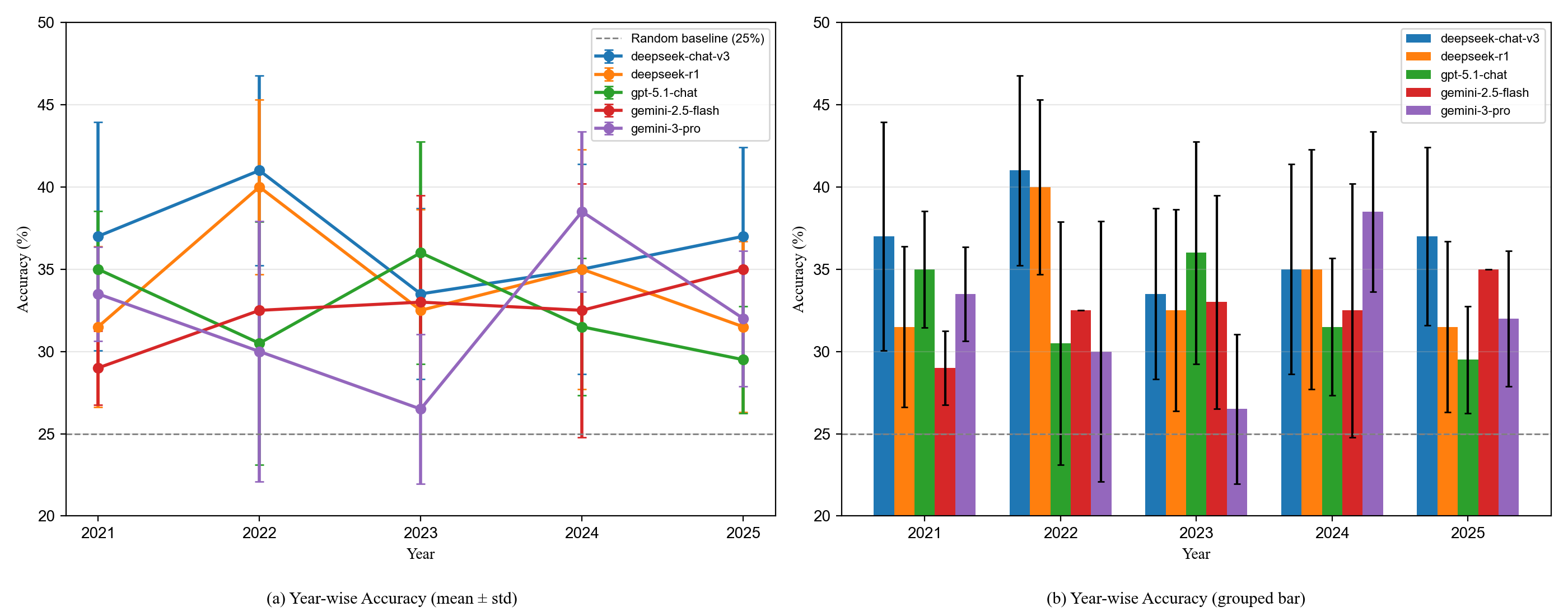}
\caption{Year-wise accuracy across models (mean $\pm$ standard deviation).}
\label{fig:yearwise_accuracy}
\end{figure*}

Performance varies non-monotonically across competition years.
Table~\ref{tab:yearwise_accuracy} reports year-wise accuracy for each model, while Figure~\ref{fig:yearwise_accuracy} visualizes the corresponding mean and dispersion across runs.
Across all models, no consistent upward or downward trend is observed, indicating that later competition sets are neither systematically easier nor harder than earlier ones.

\begin{table*}[t]
\centering
\small
\caption{Year-wise accuracy by competition year (Multi-turn).}
\label{tab:yearwise_accuracy}
\setlength{\tabcolsep}{6pt}
\begin{tabular}{lccccc}
\toprule
Model & 2021 & 2022 & 2023 & 2024 & 2025 \\
\midrule
DeepSeek-Chat-V3 & 37.0 & 41.0 & 33.5 & 35.0 & 37.0 \\
DeepSeek-R1      & 31.5 & 40.0 & 32.5 & 35.0 & 31.5 \\
GPT-5.1-Chat     & 35.0 & 30.5 & 36.0 & 31.5 & 29.5 \\
Gemini-2.5-Flash & 29.0 & 32.5 & 33.0 & 32.5 & 35.0 \\
Gemini-3-Pro     & 33.5 & 30.0 & 26.5 & 38.5 & 32.0 \\
\bottomrule
\end{tabular}
\end{table*}

The observed year-level variation reflects evolving question styles and reasoning demands in the underlying competitions.
In particular, certain years (e.g., 2022 and 2024) exhibit larger performance gaps across models, suggesting that these sets place greater emphasis on temporal composition or interaction prioritization.
This variability motivates treating each year as a distinct evaluation unit and cautions against interpreting single-year results as representative of overall model capability.

\subsection{Domain-wise Performance Profiles}
\label{subsec:domain_analysis}

We next analyze performance variation across Bazi reasoning domains under the primary multi-turn setting.
Because the original competition problems do not provide gold-standard domain or question-type annotations, we adopt a reproducible heuristic labeling scheme based on question text.
The resulting domain categories are coarse-grained and intended solely for exploratory analysis; they may contain misclassification or omission errors and should not be interpreted as definitive task taxonomies.

Table~\ref{tab:domain_accuracy} reports domain-wise accuracy aggregated over all competition years (2021--2025), computed as the ratio of correct answers to total questions within each domain.
Despite the coarse labeling, several consistent patterns emerge.

\begin{table*}[t]
\centering
\small
\caption{Domain-wise accuracy aggregated over all years (Multi-turn).}
\label{tab:domain_accuracy}
\setlength{\tabcolsep}{6pt}
\begin{tabular}{lccccc}
\toprule
Domain & DS-Chat-V3 & DS-R1 & GPT-5.1 & Gemini-2.5 & Gemini-3-Pro \\
\midrule
Career         & 27.8 & 23.3 & \textbf{31.1} & 25.6 & 23.3 \\
Health         & 32.0 & 48.0 & 20.0 & 36.0 & \textbf{60.0} \\
Family         & \textbf{53.0} & 42.0 & 38.0 & 45.0 & 40.0 \\
Other          & 29.3 & \textbf{37.1} & 22.1 & 22.9 & 24.3 \\
Education      & 22.2 & 28.9 & 26.7 & 26.7 & \textbf{31.1} \\
Personality    & \textbf{52.7} & 45.5 & 36.4 & 34.5 & 50.9 \\
Romance        & \textbf{35.1} & 27.0 & 25.9 & 28.6 & 31.9 \\
Annual fortune & 37.9 & 36.4 & \textbf{40.3} & 36.7 & 30.6 \\
Wealth         & \textbf{36.7} & 20.0 & 33.3 & 33.3 & 30.0 \\
\bottomrule
\end{tabular}
\end{table*}

Domains dominated by relatively static chart attributes, such as personality and family relations, consistently yield higher accuracy across models.
DeepSeek-Chat-V3 achieves the strongest performance in these domains, suggesting greater robustness in extracting stable symbolic relations.
In contrast, domains requiring temporal composition and interaction prioritization, particularly annual fortune, remain challenging for all models, although GPT-5.1-Chat exhibits relatively stronger performance.
Health-related questions show the largest inter-model variance, with Gemini-3-Pro and DeepSeek-R1 performing substantially better than other models.
Overall, no single model dominates across all domains, highlighting complementary strengths and the importance of domain-aware evaluation.

\subsection{Effect of Structured Reasoning Protocol}
\label{subsec:structured_reasoning}

We next examine the Structured Reasoning Protocol (SRP) as an inference-time prompting intervention.
SRP enforces an explicit ordering of reasoning steps and constrains how models attend to symbolic cues, without introducing additional domain knowledge.
Its purpose is not to improve performance per se, but to probe the sensitivity of model behavior to imposed reasoning structure under identical symbolic inputs.

\begin{figure*}[t]
\centering
\includegraphics[width=0.9\textwidth]{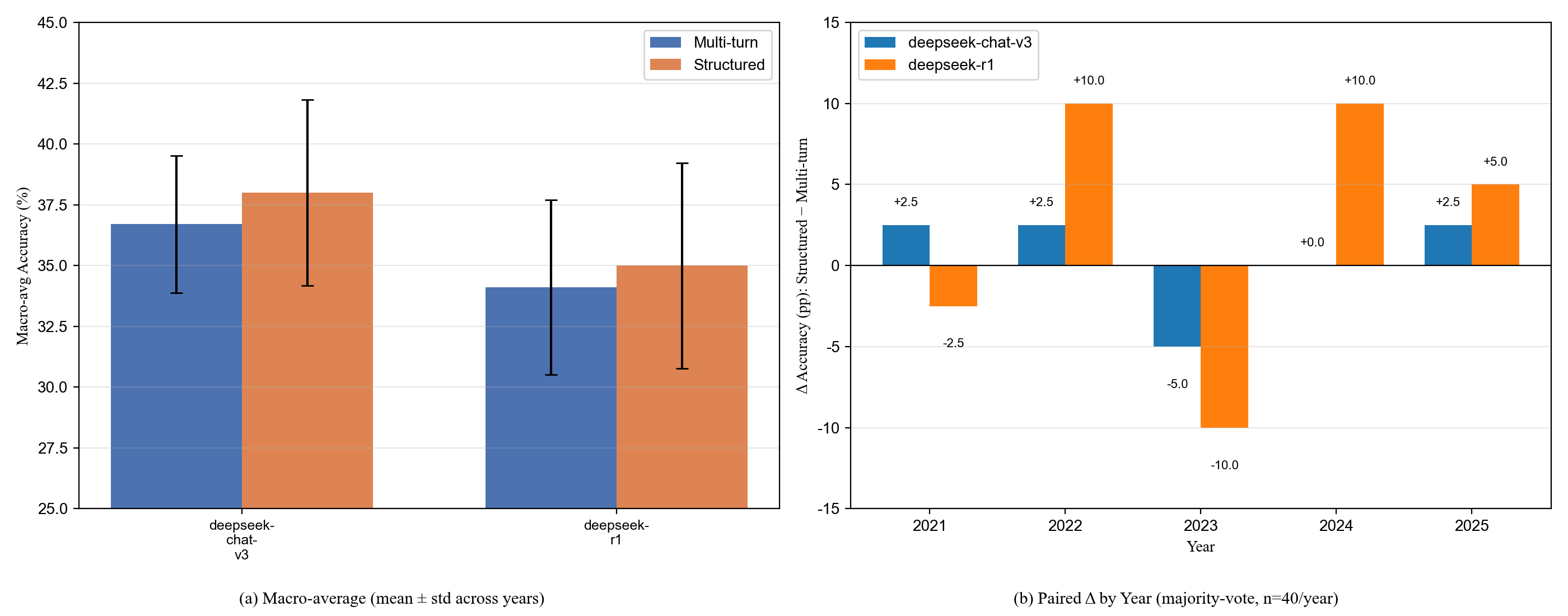}
\caption{Effect of Structured Reasoning Protocol (SRP).}
\label{fig:srp_overview}
\end{figure*}

Across models and years, SRP yields heterogeneous effects.
In some settings, it improves stability and accuracy, while in others it degrades performance.
This variability indicates that structured prompting does not provide uniform gains.
Instead, SRP reveals that model performance is sensitive to the order in which symbolic evidence is considered, particularly in questions involving competing interactions or temporal overrides.

\paragraph{Domain-specific effects on the 2025 subset.}
To further probe how reasoning structure interacts with domain characteristics, Table~\ref{tab:srp_domain_delta_2025} reports a domain-level comparison between the primary multi-turn setting and SRP for the 2025 competition set.
Because not all domains appear in 2025, this analysis is necessarily partial and should be interpreted qualitatively.

\begin{table}[t]
\centering
\small
\caption{Domain-wise accuracy change induced by Structured Reasoning Protocol (SRP) on the 2025 subset.
Values report $\Delta$ accuracy (SRP $-$ Multi-turn) in percentage points.
Positive values indicate improvement under SRP.}
\label{tab:srp_domain_delta_2025}
\setlength{\tabcolsep}{6pt}
\begin{tabular}{lcc}
\toprule
Domain & DeepSeek-Chat-V3 & DeepSeek-R1 \\
\midrule
Romance        & +2.9  & \textbf{+14.3} \\
Career         & $-$2.5 & \textbf{+15.0} \\
Education      & \textbf{+30.0} & $-$10.0 \\
Family         & $-$3.3 & $-$10.0 \\
Personality    & \textbf{+10.0} & $-$20.0 \\
Annual fortune & \textbf{+8.0}  & \textbf{+10.0} \\
Other          & \textbf{+12.0} & $-$8.0 \\
\bottomrule
\end{tabular}
\end{table}

The domain-level decomposition suggests that SRP can yield gains in domains that benefit from explicit information organization or temporal alignment (e.g., annual fortune), while reducing accuracy in domains dominated by static symbolic associations.
These mixed effects reinforce the view that structured prompting should be treated as a conditional inference-time intervention rather than a universally beneficial strategy.

\subsection{Model Agreement and Diversity}
\label{subsec:agreement}

\begin{figure}[t]
\centering
\includegraphics[width=0.95\linewidth]{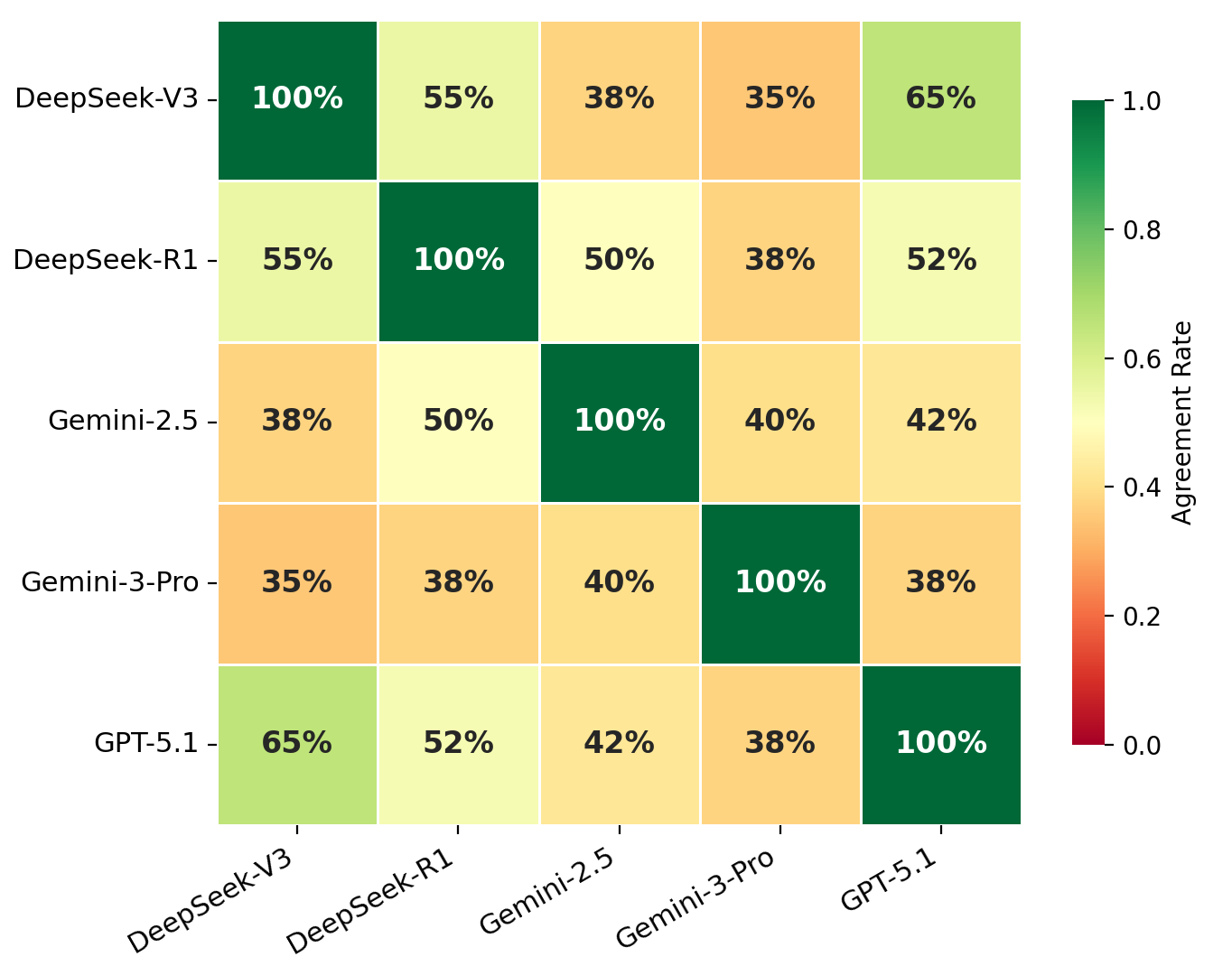}
\caption{Pairwise model agreement on the 2025 subset (Multi-turn).}
\label{fig:agreement_heatmap}
\end{figure}

Finally, we analyze pairwise answer agreement to assess behavioral similarity and potential ensemble headroom.
Figure~\ref{fig:agreement_heatmap} visualizes agreement rates between model pairs on the 2025 subset under majority-vote consolidation.

Agreement rates vary substantially, and no model pair exhibits near-perfect overlap.
This indicates that models fail on partially distinct subsets of questions, suggesting non-trivial ensemble headroom.
At the same time, moderate agreement levels imply shared failure modes, underscoring the intrinsic difficulty of temporally compositional symbolic reasoning in this domain.

\section{Discussion}
\label{sec:discussion}

This section discusses the implications of the empirical findings reported in Section~\ref{sec:evaluation}.
The goal is not to re-state quantitative results, but to interpret what the observed patterns suggest about the nature of symbolic reasoning in \textsc{BaziQA-Benchmark},
the behavior of contemporary large language models under such conditions, and the design of future evaluation benchmarks.

\subsection{What Does the Benchmark Measure?}

The results indicate that \textsc{BaziQA-Benchmark} captures a form of reasoning that is neither purely factual recall nor unconstrained text generation.
Performance above the random baseline across all evaluated models confirms that the benchmark contains learnable structure.
At the same time, absolute accuracy remains far from saturation, suggesting that this structure is not trivially exploitable through surface heuristics alone.

Unlike many existing cultural or metaphysical evaluations, the benchmark emphasizes discrete decision-making under explicit symbolic constraints.
Each question requires the model to select one option from a fixed set, often under the interaction of multiple symbolic factors and temporal layers.
This design isolates reasoning correctness from stylistic fluency and verbosity, and therefore provides a more controlled probe of symbolic inference behavior.

\subsection{Temporal Composition as a Core Difficulty}

One of the most consistent observations across models is substantial variation in performance across competition years.
The absence of a monotonic trend suggests that year-wise difficulty is driven more by changes in reasoning structure than by incremental dataset difficulty.

In particular, several years place stronger emphasis on temporal composition, requiring the integration of natal chart information with decadal and annual cycles.
Such questions demand not only recognition of relevant symbolic elements, but also their prioritization under temporal context.
The observed sensitivity to year-level variation indicates that temporal composition remains a major challenge for current models.

This finding aligns with broader observations in symbolic and multi-step reasoning benchmarks, where models often struggle when multiple dependencies must be evaluated jointly rather than independently. \textsc{BaziQA-Benchmark} provides a naturally occurring instance of this difficulty, grounded in a real-world symbolic system.

\subsection{Domain Heterogeneity and Reasoning Modes}

Domain-wise analysis reveals substantial heterogeneity in model performance.
Domains dominated by static chart attributes, such as personality and family relations, tend to exhibit higher accuracy across models.
In contrast, domains involving temporal change or contingent outcomes, such as annual fortune, remain challenging even for the strongest models.

Importantly, no single model dominates across all domains.
This suggests that different model architectures or training regimes may favor different reasoning modes, such as static pattern recognition versus temporally conditioned inference.
From a benchmark design perspective, this heterogeneity is desirable, as it reduces the likelihood that overall performance is driven by a narrow subset of question types.

The observed domain imbalance further highlights the importance of reporting domain distributions alongside domain-wise accuracy.
Without such context, domain-level comparisons may be misleading, particularly for sparsely represented categories.

\subsection{Interpreting the Effect of Structured Reasoning}

The Structured Reasoning Protocol (SRP) provides insight into how explicit reasoning constraints affect model behavior.
The results show that SRP does not uniformly improve accuracy across models or years. Instead, its effect is heterogeneous and context-dependent.

At the macro level, SRP yields modest gains and reduced variance for some models, indicating more stable behavior across different evaluation conditions.
However, year-wise paired analysis reveals both positive and negative effects, suggesting that enforcing a fixed reasoning order can sometimes misalign with the structure of specific questions.

These findings caution against interpreting structured prompting as a general-purpose solution for symbolic reasoning.
Rather than serving as a performance enhancement, SRP functions more effectively as a diagnostic tool.
By constraining reasoning order, it exposes how models prioritize symbolic evidence and how sensitive they are to interaction ordering.

From an evaluation standpoint, this highlights the value of protocol-based analysis for understanding model behavior beyond aggregate accuracy.

\subsection{Systematic Failures and Model Diversity}

The presence of hard items—questions that no evaluated model answers correctly—points to systematic limitations shared across architectures.
Many of these items require precise temporal localization or the simultaneous satisfaction of multiple symbolic conditions.
Error probability compounds across such sub-inferences, making joint correctness difficult even when individual components are moderately well handled.

At the same time, inter-model agreement analysis reveals that models do not fail on identical subsets of questions.
This diversity suggests that errors arise from different inductive biases rather than from a single shared weakness.
While this observation motivates potential ensemble approaches, it also underscores the importance of analyzing disagreement patterns when interpreting benchmark results.

\subsection{Implications for Benchmark Design}

Taken together, these observations suggest several considerations for future symbolic reasoning benchmarks.
First, temporal composition should be treated as a first-class evaluation dimension, rather than an incidental feature.
Second, domain heterogeneity should be preserved and explicitly reported to avoid overfitting to narrow reasoning modes.
Third, evaluation protocols themselves can serve as analytical instruments, revealing model sensitivities that are not apparent from single-shot accuracy.

\textsc{BaziQA-Benchmark} illustrates how culturally grounded symbolic systems can be used to construct rigorous and reproducible evaluation suites without relying on synthetic task design.
This approach complements existing benchmarks by introducing naturally structured, non-standard reasoning environments.

\section{Conclusion}
\label{sec:conclusion}

This paper introduces \textsc{BaziQA-Benchmark}, a standardized evaluation suite for assessing large language models on structured symbolic reasoning grounded in Bazi astrology.
The benchmark is constructed from professionally curated competition problems and supports objective scoring under controlled evaluation protocols.

Through extensive empirical analysis, we show that contemporary models exhibit non-trivial but limited capability on this benchmark.
Performance varies substantially across years and domains, remains sensitive to temporal composition, and is influenced by reasoning protocol design.
These properties distinguish the benchmark from anecdotal or prompt-driven evaluations and position it as a diagnostic tool for symbolic and temporally compositional reasoning.

Beyond reporting accuracy, this work emphasizes the importance of evaluation design itself.
By combining competition-grade data, multi-turn interaction, protocol-based analysis, and detailed diagnostics, \textsc{BaziQA-Benchmark} provides a framework for probing how models reason under structured, non-standard symbolic systems.

We hope this benchmark will support future research on symbolic reasoning, temporal inference, and evaluation methodology, and encourage the exploration of diverse knowledge systems as rigorous testbeds for language model assessment.

\bibliographystyle{IEEEtran}
\bibliography{references}

\end{document}